\documentclass[twoside,twocolumn]{article}
\usepackage{graphicx}		 % Need this to include images
\usepackage{hyperref}
\usepackage{amsmath}
\usepackage{amssymb}
\usepackage[]{algorithm2e}
\usepackage[section]{placeins}
\graphicspath{ {images/} }

\usepackage[
backend=biber,
style=numeric,
sorting=none
]{biblatex}
\usepackage[english]{babel}
\usepackage[utf8]{inputenc}
\title{Neural Network Augmented Compartmental Pandemic Models}
\author{Lorenz Kummer, Kevin Sidak}
\date{July 2021}

\begin{document}

\maketitle

\begin{abstract}
    Compartmental models are a tool commonly used in epidemiology for the mathematical modelling of the spread of infectious diseases, with their most popular representative being the Susceptible-Infected-Removed (SIR) model and its derivatives. However, current SIR models are bounded in their capabilities to model government policies in the form of non-pharmaceutical interventions (NPIs) and weather effects and offer limited predictive power. More capable alternatives such as agent based models (ABMs) are computationally expensive and require specialized hardware. We introduce a neural network augmented SIR model that can be run on commodity hardware, takes NPIs and weather effects into account and offers improved predictive power as well as counterfactual analysis capabilities. We demonstrate our models improvement of the state-of-the-art modeling COVID-19 in Austria during the 03.2020 to 03.2021 period and provide an outlook for the future up to 01.2024.

\end{abstract}
\section{Introduction}
\label{sec:intro}

In the period between March 15th 2020 and  April 4th, Austria enacted it’s first series of COVID-19 NPIs \cite{cov19law, cov19law2, cov19law3, cov19law4, cov19law5, cov19law6, cov19law7, cov19law8, cov19law18, cov19law19, cov19law20}, resulting in a country-wide lockdown and restrictions of public life, followed by a series of repeated loosening and tightening regulations that have not been completely taken back until today \cite{cov19law10, cov19law11, cov19law12, cov19law13, cov19law14, cov19law15, cov19law16, cov19law17}. Nonetheless, as of March 1st 2021, 8574 people were registered as deceased with or by COVID-19 \cite{opendata21covid, jhu21covid} in Austria. Although a highly socio-politically relevant question, it remains unclear how effective these NPIs were, what would have happened without any NPIs or only a subset of the historical NPIs and whether a less intrusive set of NPIs exists which provides an outcome similar to the historical one. These are all counterfactual questions which cannot be empirically answered per definition but only through simulation.
Given the correlation between weather and pandemic spread \cite{ma2020weather, xi2020weather, sahin2020weather, tosepu2020weather}, we recognize that any such simulation must not only model the pandemic given a certain set of NPIs, but also account for weather effects. 
\subsection{Related Work}
\label{ssec:related}
ABMs, used by Austrian government agencies as decision support tool \cite{bmsgpk21covidmodels}, are powerful instruments for simulating disease spread under NPIs (e.g. contact tracing \cite{bicher2021evaluation}) as they simulate individual agents and their movement, behaviour and interactions but are also capable of evaluating immunization levels \cite{bicher2021model} and vaccination strategies \cite{jahn2021targeted} as well as estimating undetected cases \cite{rippinger2021evaluation}. However, they are computationally complex \cite{bicher2021model} which limits their use to institutions with access to sufficient computing power or specialized hardware \cite{kosiachenko2019mass}.
\\\\
A computationally less expensive model is the SIR model. Since its inception \cite{kermack1927contribution}, the SIR model has been extended numerous times \cite{li2009generalization, chen2020time, acemoglu2020multi} and it recently found an application in the modelling of COVID-19 \cite{cooper2020sir, covid2021ihme, murray2021potential, bmsgpk21covidmodels, ndiaye2020analysis, munoz2021sir}. The predictive power of the SIR model in the context of COVID-19 and been confirmed by \cite{friedman2021predictive}. Currently though, SIR models are limited in their capability to simulate NPIs or weather effects, with proposed solutions ranging from segmenting and picking epidemiological parameters for various phases of the pandemic manually or automatic change point detection and inference of epidemiological parameters for different phases of the pandemic \cite{covsirphy} to training simple exponential regression models on predicting epidemiological parameters \cite{munoz2021sir}. Other publications use projections of the effects of specific NPIs on reproduction number based on empirical studies of that particular NPI and incorporate them in the SIR-Model, but completely disregard weather effects \cite{covid2021ihme}. In any case, the mix of varying weather conditions and large combinations of different NPIs can not be modelled by SIR models currently.
\subsection{Motivation and Contribution}
\label{ssec:mot}
We propose separating the problem of estimating the epidemiological parameters as function of weather conditions and NPIs, which we do in an inductively learnt pre-prepossessing step, from the computation of the SIR model and to apply an inductively learnt post-processing step to the SIR-models result to compensate for the SIR-models intrinsic lack of representative power. Using our technique, we achieve improved predictive power at slightly higher computational cost than running non-augmented SIR models and are able to provide an efficient and accurate tool for backwards analysis of NPIs.

\section{Model Pipeline}
\label{sec:pipe}
\begin{figure}%[!h]
\includegraphics[width=0.5\textwidth]{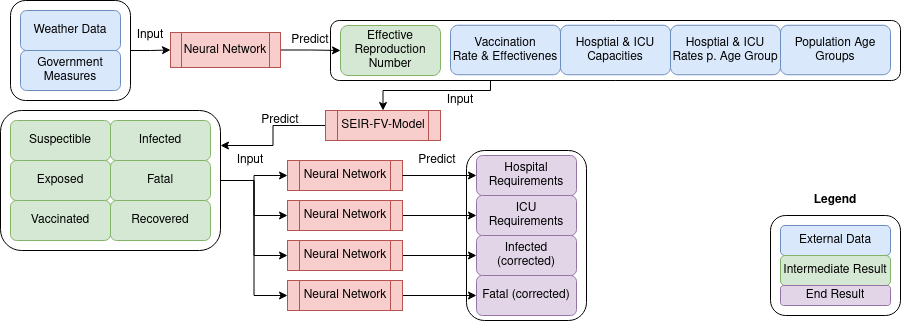}
\caption{\label{fig:pipe}\textit{Model Pipeline}}
\end{figure}

As mentioned in sec. \ref{ssec:mot}, we separated the modelling of the pandemic in three separate tasks. Our pipeline, as illustrated in fig. \ref{fig:pipe}, consists modelling reproduction number as function of weather and NPIs, which we accomplish using a neural network, modelling the spread of COVID-19 based on the modelled reproduction number using a SEIR-FV-Model and subsequently using the SEIR-FV-Model output as input for a group of neural networks correcting SEIR-FVs intrinsic error and inferring hospital resource utilization.
\subsection{Reproduction Number} 
\label{ssec:repr}
For modelling reproduction number $R_{eff}$ as function of weather and NPIs, we used numerically encoded NPI data from Oxford COVID-19 Government Response Tracker (OxGRT) \cite{hale2021global} which is already provided in machine-learnable format at daily resolution, monthly weather data from 1955 till today provided by Hohe Warte weather station near Austria's capital Vienna \cite{hohewarte} and daily reproduction number estimates provided by AGES \cite{agesReff}. Zentralanstalt für Meteorologie und Geodynamik (ZAMG) \cite{zamg}, which could provide more fine granular weather data not only for Vienna but all of Austria, was contacted but unfortunately did not wish to cooperate. In order to be suitable input for a neural network, weather data is first up sampled to match OxGRT datas resolution using Fouriers method and most significant features are selected using linear regression-analysis. Based on the selected features, additional features are engineered via power and log transforms to allow more room for choice in the forthcoming step. Next OxGRT data and weather data are merged and scaled and linear regression-analysis is employed again to eliminate independent variables not affecting the dependent variable $R_{eff}$ for a significance level of $p=0.05$ with exception to NPIs. For NPIs, we manually selected which to keep and which to remove in order to be able to keep the model from picking measures not directly relevant for pandemic spread such as "Fiscal Measures", "Debt Contract Relief" or "Income Support" but nonetheless part of the OxGRT dataset and because we wanted measures with particularly severe potential human collateral damage (e.g. "School Closures", "Workplace Closing") (CITE) to be kept even if not chosen for the given significance level. Finally, after the preprocessing of the training data is finished, we train a neural network on each of the 3 $R_eff$ estimates (lower, upper, mean) provided by AGES. We chose a neural network with non-linear activation functions over a linear regression model because despite achieving a good fit (R2=0.99) during regression analysis in previous steps, the linear regression model performed very poorly for OOS data (we found setting multiple NPIs in the dataset to 0.0 could lead to the linear regression model predicting negative reproduction numbers), a deficit we did not find using a non-linear neural network. 
\begin{figure}%[!h]
\includegraphics[width=0.5\textwidth]{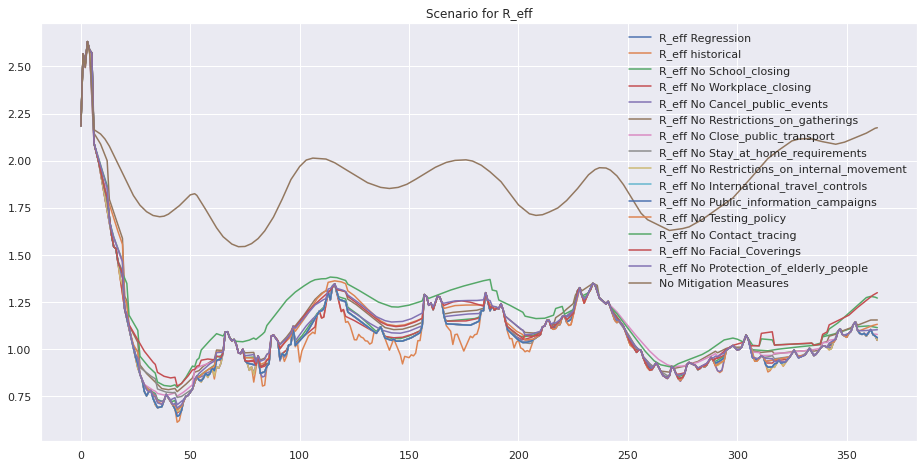}
\caption{\label{fig:seirfv}\textit{Predicting $R_{eff}$}}
\end{figure}
\\\\
Dependent on application, for backwards-analysis we apply a post-processing step that introduces historical $R_{eff}$ estimates as lower bound for scenarios where a specific NPI is completely removed s.t. e.g. in a scenario without mandatory face masks, $R_{eff}$ can never be lower than in a historical case where face masks were mandatory. For forward-analysis the post-processing step is naturally not applied.
\begin{table}%[!h]
\centering
\begin{tabular}{|l|lll|}
\hline
\textbf{} & \textbf{Low} $R_{eff}$ & \textbf{Mean $R_{eff}$} & \textbf{High $R_{eff}$} \\ \hline
\textbf{R2}  & 0.94 & 0.93 & 0.94 \\ \hline
\end{tabular}
\caption{\label{tab:overview1} R2 scores for predicting $R_{eff}$ on 30\% validation set via a neural network}
\end{table}

\subsection{Pandemic Model}
\label{ssec:pand}
\begin{figure}%[!h]
\includegraphics[width=0.5\textwidth]{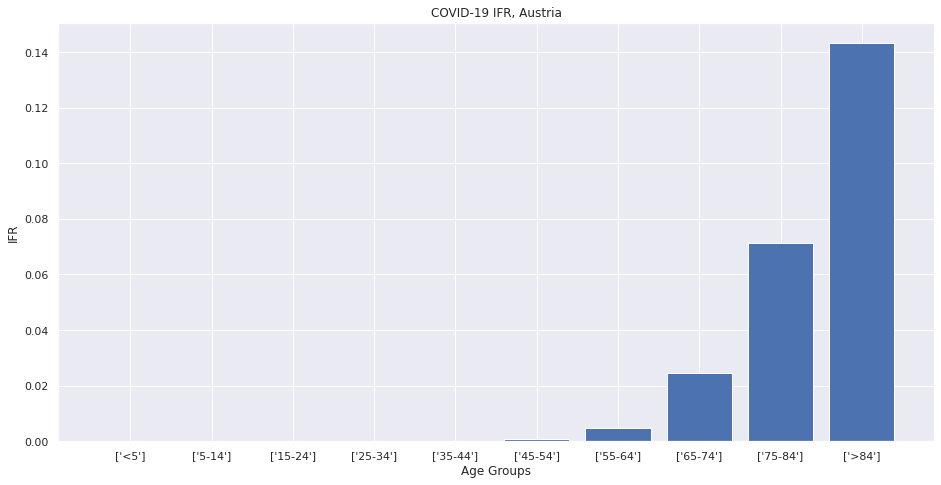}
\caption{\label{fig:ifr}\textit{Per Age Group IFR}}
\end{figure}

\begin{figure}%[!h]
\includegraphics[width=0.5\textwidth]{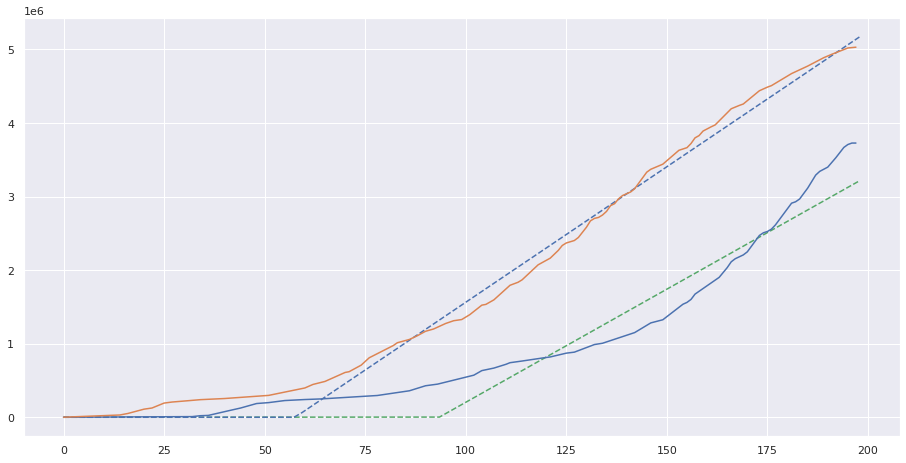}
\caption{\label{fig:vac}\textit{Vaccination Rate in Austria}}
\end{figure}

The predicted lower, mid and upper vectors containing daily $R_{eff}$ values for a given time interval is then handed to three SEIR-FV models (one for each estimate of lower, mid, upper $R_{eff}$) which, for each age group, tracks \textbf{S}usceptible, \textbf{E}xposed, \textbf{I}nfectious, \textbf{R}ecovered, \textbf{F}atal, \textbf{V}accinated population groups, see fig. \ref{fig:seirfv} for a simplified illustration. 
\begin{figure}%[!h]
\includegraphics[width=0.5\textwidth]{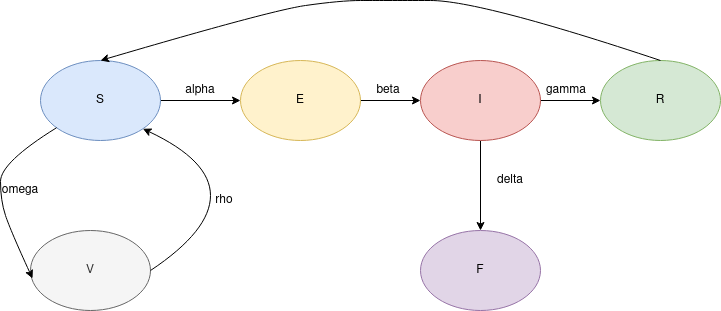}
\caption{\label{fig:seirfv}\textit{\textbf{S}usceptible, \textbf{E}xposed, \textbf{I}nfectious, \textbf{R}ecovered, \textbf{F}atal, \textbf{V}accinated }}
\end{figure}
The initial age group distribution is obtained from \cite{austat}, Hospitalization, Intensive Care Unit (ICU), Infection Fatality (IFR) (fig. \ref{fig:ifr}) ratios per age group is obtained from \cite{agesDash}. 
The age groups data is homogenized to groups from $<5$ to $>84$ in 5-year intervals. The rate at which people are vaccinated with any vaccine is obtained from \cite{covid19datahub} and for ease of use, an average daily vaccine doses available per day is computed by fitting a linear function through the data (fig. \ref{fig:vac}) and using its slope. 
The SEIR-FV models follow an old-first vaccination strategy, i.e. available vaccine doses are distributed among age groups, preferring older age groups over younger age groups. For duration of vaccine and recovery immunity and efficiency, we use estimates based on \cite{ripperger2020orthogonal}.
Incorporating ICU and Hospitalization rates into the model is done statistically: based on the a-priory known hospitalization and ICU rates for each age group as well as the current number of infected computed by the SEIR-FV model, we infer the number of persons of an age group requiring a hospital bed or an ICU bed. For cases where hospital requirements exceed hospital resources, we made the simplifying assumption that all patients requiring hospital treatment but not receiving it become a fatality. 
\subsection{Error Correction Networks}
\label{ssec:errcorrnn}
\begin{figure}[!h]
\includegraphics[width=0.5\textwidth]{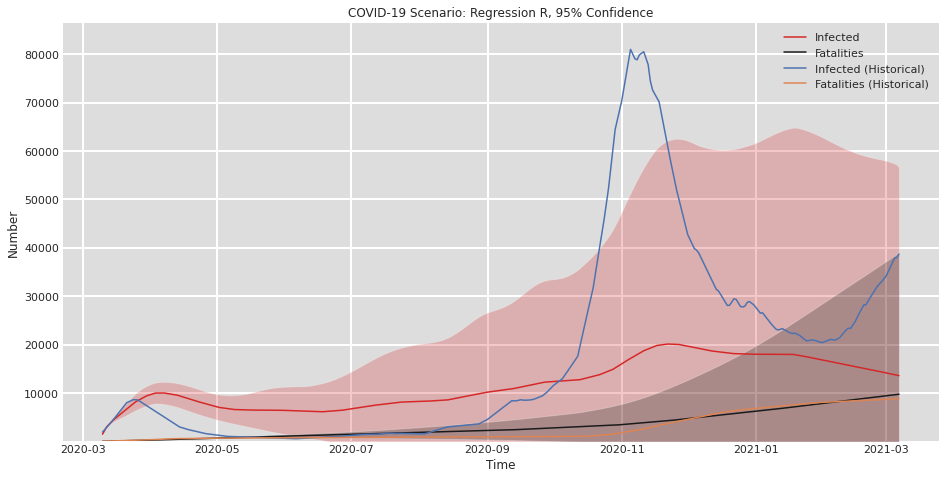}
\caption{\label{fig:regr}\textit{SEIR-FV, R obtained from NN Regressor}}
\end{figure}

\begin{figure}[!h]
\includegraphics[width=0.5\textwidth]{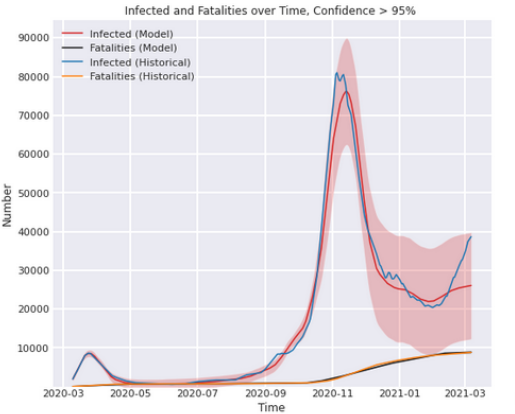}
\caption{\label{fig:errcorrnn}\textit{NN Error Correction Augmented SEIR-FV, R obtained from NN Regressor, 03.2020 to 03.2021 time frame}}
\end{figure}

\begin{figure}%[!h]
\includegraphics[width=0.5\textwidth]{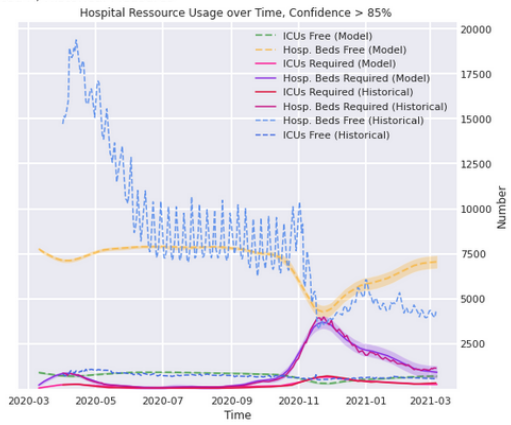}
\caption{\label{fig:hospnn}\textit{Hospital Resource Requirements computed by NN, 03.2020 to 03.2021 time frame}}
\end{figure}

The pandemic models provided by SEIR-FV even for the time dependent $R_{eff}$ computed from weather and NPI data by the NN regressor is, while qualitatively making accurate predictions where the peaks in disease spread are, quantitatively off by orders of magnitude (fig. \ref{fig:regr}). We solve this by training so-called error correction neural networks (see sec. \ref{sec:pipe}, fig. \ref{fig:pipe} for the step in the modelling pipeline) on correcting the SEIR-FVs outputs error (fig. \ref{fig:errcorrnn}) and furthermore predict the amount of hospital resources used (fig. \ref{fig:hospnn}).
As illustrated by the figures, our error correction neural networks approach greatly improves the SEIR-FVs outputs quantitatively, with the historical numbers of infected and fatal cases being inside the given 95\% confidence interval for the whole 03.2020 to 03.2021 time period.
For hospital resources (regular and ICU beds), the augmented model produces a quantitatively and qualitatively equally accurate output for a 85\% confidence interval. Regarding network architecture, we use simple 4-Layer ReLU activated MLPs. 
To account for the uncertainty that comes with using such networks that always incorporate randomness due to their random initial state, we train groups of 10 networks for each of the target variables (infected, fatal, hospital bes, ICU) and provide the results in the form of confidence intervals. We do this for each of the three SEIR-FV models computed for the three estimates for $R_{eff}$.
\subsection{NPI Collateral Damage Model}
\label{ssec:coldammod}
\begin{figure}%[!h]
\includegraphics[width=0.5\textwidth]{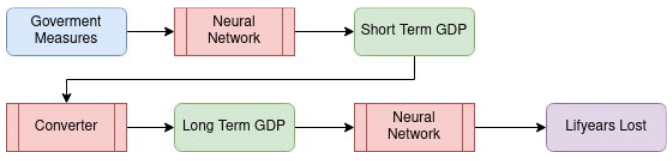}
\caption{\label{fig:coldammod}\textit{NPI Collateral Damage Model}}
\end{figure}

Secondary to modelling pandemic spread, we decided to model human collateral damage to NPIs as well. Our simple collateral damage model exploits the known correlation between life expectancy and health and economic development \cite{stuckler2009public, tapia2017population, reeves2015economic} and the observed economic depression \cite{steiber2021covid, holzl2021zweite} during the pandemic, which we conjecture to be caused by business and border closures. Other potential causes of collateral damage such as mental health \cite{pieh2020effect, pieh2021comparing, pieh2021mental, stieger2020psychological, stieger2021emotional, stolz2021impact, ueda2020suicide, radeloff2021trends, ruck2020will, nowotny2019depressionsbericht, alboni2008there, jackson2013depression}, disrupted medical supply chains \cite{freudenberg2020impact}, skipping of early medical examinations \cite{metzler2020decline} and problems for people with disabilities \cite{negrini2020up} were initially considered, but we decided to disregard them due to a lack of useful coherent data. For our economy-based collateral damage model, we first confirmed that a statistically significant (p=0.05) relation between economic factors and life expectancy existed in pre-corona times, a purpose for which we used OECDs GDP and life-years-lost (LYL) data sets \cite{oecd}, and that the same holds for NPIs and GDP. Next, as depicted in fig. \ref{fig:coldammod} we trained a NN to predict GDP dependent on NPIs and another NN to predict LYL dependent on GDP to finally obtain a pipeline modelling LYL dependent on NPIs. Admittedly, this approach suffers from not taking the collateral damage produced by corona itself into account (which would certainly be strong in the case of e.g. an unrestricted spread of the pandemic) but the lack of a baseline for such scenarios made it impossible to model them.

\section{Results}
We applied our approach in two different tasks. First, we applied it to the backwards analysis task, i.e. asking the counterfactual question "What would have happened if NPI $<x>$ had not been in place?" to find out which NPIs were most crucial in the past. Second, we apply our approach to predicting future developments. In this case, we will make a simple statistical weather prediction for future weather based on the historical data provided by Hohe Warte weather station to use as input for our NN regressors trained to predict $R_{eff}$.
% iso accuracy: marvin erreicht iso accuracy mit X% des aufwandes
% herausarbeiten
% trennlinie im plot
% aufwand für den fwd pass in abhängigkeit von bitbreite analystisch. einfaches modell. cool wärs alles in eine zahl zusammenzufasen, wieviel etwas kostet. mann kan auch sagen, dass fp16 doppelt so schnell ist wie fp32 als einfachstes modell. Ansatz lineare aufwand zwischen aufwand.
\subsection{Backwards Analysis}
\begin{figure}%[!h]
\includegraphics[width=0.5\textwidth]{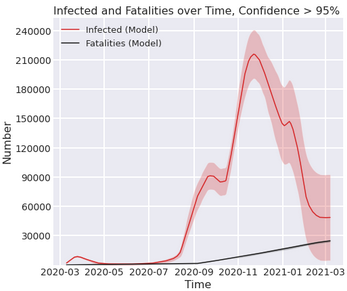}
\caption{\label{fig:no_facial}\textit{No Facial Masks, Infected and Fatalities, 03.2020 to 03.2021 time frame}}
\end{figure}

\begin{figure}%[!h]
\includegraphics[width=0.5\textwidth]{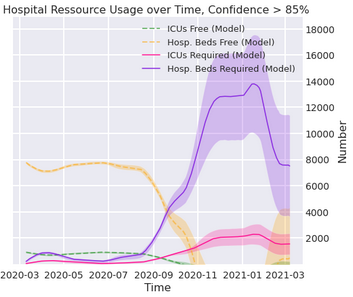}
\caption{\label{fig:no_facial_hosp}\textit{No Facial Masks, Hospital Ressource Requirements computed by NN, 03.2020 to 03.2021 time frame}}
\end{figure}

\begin{figure}%[!h]
\includegraphics[width=0.5\textwidth]{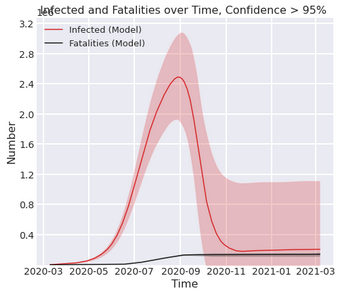}
\caption{\label{fig:no_mitis}\textit{No Mitigation Measures, Infected and Fatalities, 03.2020 to 03.2021 time frame}}
\end{figure}

\begin{figure}%[!h]
\includegraphics[width=0.5\textwidth]{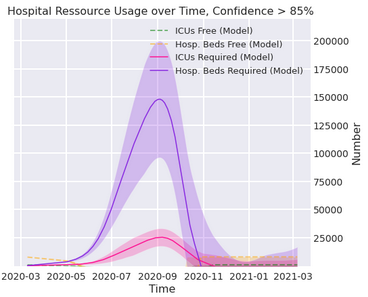}
\caption{\label{fig:no_mitis_hosp}\textit{No Mitigation Measures, Hospital Resource Requirements computed by NN, 03.2020 to 03.2021 time frame}}
\end{figure}

For backwards analysis of NPI effectiveness, we follow a "take one out" approach. We compute each scenario under the assumption of historical weather and NPIs but with a particular NPI not implemented, i.e. it's zeroed in the input data matrix. 
\begin{figure}%[!h]
\includegraphics[width=0.5\textwidth]{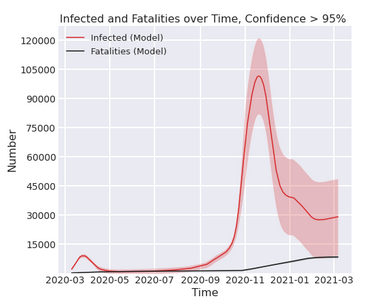}
\caption{\label{fig:no_wp}\textit{No Workplace Closing, Infected and Fatalities, 03.2020 to 03.2021 time frame}}
\end{figure}
\begin{figure}%[!h]
\includegraphics[width=0.5\textwidth]{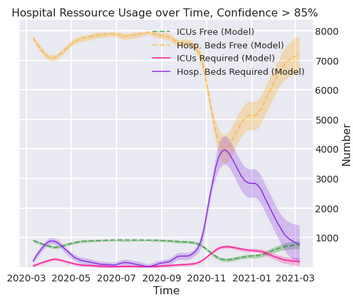}
\caption{\label{fig:no_wp_hosp}\textit{No Workplace Closing, Hospital Ressource Requirements computed by NN, 03.2020 to 03.2021 time frame}}
\end{figure}
Fig. \ref{fig:no_wp}, \ref{fig:no_wp_hosp} display infections, fatalities and hospital resource usage for the hypothetical scenario where no workplaces closures were implemented. The model predicts a slightly higher peak during the winter months and accordingly a higher hospital resource consumption. Particularly critical is the elongated period ICUs are over capacity compared to the historical scenario, leading to slightly higher number of fatalities in vulnerable age groups. Another potentially interesting scenario is no obligatory facial mask usage (fig. \ref{fig:no_facial}, \ref{fig:no_facial_hosp}). Infections and fatalities nearly triple for this scenario, hospital resources are catastrophically overburdened for a period of over six months which contributes massively to fatalities in older population groups likelier to require ICU care in case of an infection.
Finally, we want to explore the case where no mitigation measures were implemented at all and COVID-19s spread is only controlled by weather dynamics and population composition (fig. \ref{fig:no_mitis}, \ref{fig:no_mitis_hosp}). In this scenario, we see a massive peak for infections in September, with multiple million people suffering COVID-19 infections simultaneously. The massive and constant operation of hospitals over capacity over the whole evaluated time frame leads to up to a quarter million fatalities within the first 12 months of the pandemic.
\subsubsection{NPI Collateral Damage}
\label{sssec:coldam}
\begin{figure}%[!h]
\includegraphics[width=0.5\textwidth]{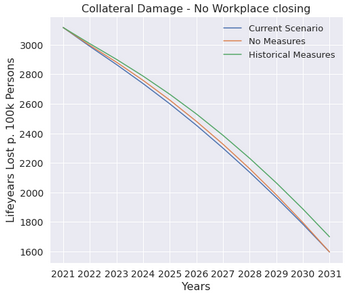}
\caption{\label{fig:no_wp_closing_col}\textit{No Workplace Closing, Collateral Damage, 03.2020 to 03.2021 time frame}}
\end{figure}

Using the collateral damage model described in sec. \ref{fig:coldammod}, computed collateral damage for two scenarios to illustrate the functionality of our model. Fig. \ref{fig:no_wp_closing_col} illustrates the scenario where historical NPIs were in place, except "Workplace Closing". As shown by the visualization, removing the "Workplace Closing" NPI dramatically reduces collateral damage, with the prediction of LYL for the next decade being not worse than what would have been the case without any NPI induces collateral damage.
A scenario with historical NPIs but no facial masks is shown by fig. \ref{fig:no_wp_closing_col}. As you can see, our collateral damage model predicts only a slight improvement compared to historical measures.
\begin{figure}%[!h]
\includegraphics[width=0.5\textwidth]{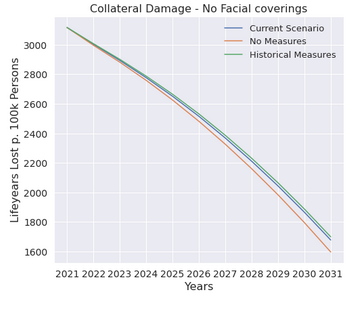}
\caption{\label{fig:no_facial_col}\textit{No Facial Masks, Collateral Damage, 03.2020 to 03.2021 time frame}}
\end{figure}

\subsection{Predictive Power}
\begin{figure}%[!h]
\includegraphics[width=0.5\textwidth]{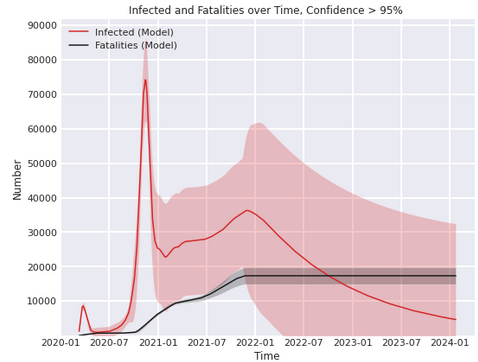}
\caption{\label{fig:pred_long}\textit{Long Term Prediction computed by NN, 03.2020 to 03.2024 time frame}}
\end{figure}

\begin{figure}%[!h]
\includegraphics[width=0.5\textwidth]{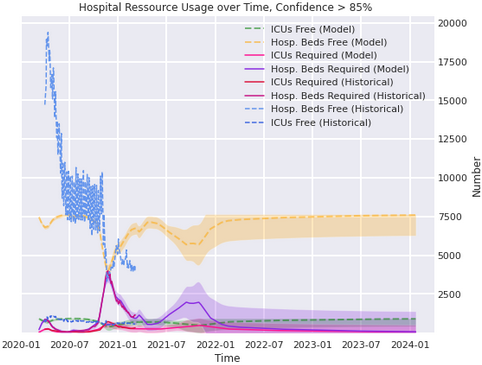}
\caption{\label{fig:pred_long_hosp}\textit{Long Term Hospital Requirements computed by NN, 03.2020 to 03.2024 time frame}}
\end{figure}

For long-term predictions illustrated in fig. \ref{fig:pred_long} and \ref{fig:pred_long_hosp} the model pipeline is fitted on historical data of the 03.2020 to 03.2021. Future weather data for the 03.2021 to 01.2024 period is statistically generated from historical weather data and government measures are assumed to be repeated on a yearly basis, but of course scenarios for alternative sets of future government measures could be computed as well. As you can see, the model predicts with a high probability a potential future peak in infections as well as ICU requirements exceeding capacity in December 2021 and January 2022. After the winter 2021/2022 peak, the number of infected slowly declines and fatalities have reached their maximum due to vulnerable age groups having been immunized via recovery or vaccination.
\\\\
The predictive power of the model pipeline is to be taken with a grain of salt, though. The current prediction algorithm is only capable to predict future developments from the very beginning of the pandemic due to the way the SEIR-FV model is initialized. Correcting this deficit will be subject to future work, but it offsets the predictions quantitatively.
\section{Conclusion}
We present a new approach of pandemic modelling through augmenting a regular SEIR-FV compartmental pandemic model with pre- and post-processing neural networks prediction reproduction number dependent on weather conditions and NPIs and demonstrate its usefulness for backwards analysis tasks as well its potential for predicting future developments w.r.t. COVID-19. Admittedly, the predictive power of our model leaves room for improvements, but we think that despite our approaches deficits, it still advances the state-of-the-art in cheap but accurate compartmental pandemic models due to its ability to incorporate weather data and fine granular NPI data from OxGRT. We also provide a simple NPI collateral damage model capable of predicting the development of future life-years-lost dependent on NPIs.

\section{Future Work}
We intend to fix the pandemic model's current issues when making long-term predictions and release it in the form of a web-based interface to allow the public to play out hypothetical backwards analysis scenarios for various sets of NPIs as well as to make future predictions dependent on user chosen NPI sets and weather conditions. Furthermore, we intend to use this report as a first draft for a publication in a journal or a conference focusing on pandemic modelling. The web interface and the potential journal or conference paper will not contain the collateral damage model because we consider it too weak to support the potentially strong political implications that could be derived from it.
%Bei conference papers oft keine future work. Wenn platzprobleme, entfernen wir die section und bauen das als subsection in conclusion.
% Jitter von push up wegen loockback über gradient diversitry. 2nd order Hessians?
\subsection{}

\printbibliography

@article{kermack1927contribution,
  title={A contribution to the mathematical theory of epidemics},
  author={Kermack, William Ogilvy and McKendrick, Anderson G},
  journal={Proceedings of the royal society of london. Series A, Containing papers of a mathematical and physical character},
  volume={115},
  number={772},
  pages={700--721},
  year={1927},
  publisher={The Royal Society London}
}

@article{li2009generalization,
  title={Generalization of the Kermack-McKendrick SIR model to a patchy environment for a disease with latency},
  author={Li, J and Zou, X},
  journal={Mathematical Modelling of Natural Phenomena},
  volume={4},
  number={2},
  pages={92--118},
  year={2009},
  publisher={EDP Sciences}
}

@article{chen2020time,
  title={A time-dependent SIR model for COVID-19 with undetectable infected persons},
  author={Chen, Yi-Cheng and Lu, Ping-En and Chang, Cheng-Shang and Liu, Tzu-Hsuan},
  journal={IEEE Transactions on Network Science and Engineering},
  volume={7},
  number={4},
  pages={3279--3294},
  year={2020},
  publisher={IEEE}
}

@article{munoz2021sir,
  title={A SIR-type model describing the successive waves of COVID-19},
  author={Mu{\~n}oz-Fern{\'a}ndez, Gustavo A and Seoane, Jes{\'u}s M and Seoane-Sep{\'u}lveda, Juan B},
  journal={Chaos, Solitons \& Fractals},
  volume={144},
  pages={110682},
  year={2021},
  publisher={Elsevier}
}

@book{acemoglu2020multi,
  title={A multi-risk SIR model with optimally targeted lockdown},
  author={Acemoglu, Daron and Chernozhukov, Victor and Werning, Iv{\'a}n and Whinston, Michael D and others},
  volume={2020},
  year={2020},
  publisher={National Bureau of Economic Research Cambridge, MA}
}

@misc{covsirphy,
  title = {COVID-19 analysis with phase dependent SIRs},
  howpublished = {\url{https://lisphilar.github.io/covid19-sir/}},
  note = {Accessed: 2021-07-12}
}

@article{covid2021ihme,
author = {Reiner, Robert and Barber, Ryan and Collins, James and Zheng, Peng and Adolph, Christopher and Albright, James and Antony, Catherine and Aravkin, Aleksandr and Bachmeier, Steven and Bang-Jensen, Bree and Bannick, Marlena and Bloom, Sabina and Carter, Austin and Castro, Emma and Causey, Kate and Chakrabarti, Suman and Charlson, Fiona and Cogen, Rebecca and Combs, Emily and Murray, Christopher},
year = {2021},
month = {01},
pages = {},
title = {Modeling COVID-19 scenarios for the United States},
volume = {27},
journal = {Nature Medicine},
doi = {10.1038/s41591-020-1132-9}
}

@article{cooper2020sir,
  title={A SIR model assumption for the spread of COVID-19 in different communities},
  author={Cooper, Ian and Mondal, Argha and Antonopoulos, Chris G},
  journal={Chaos, Solitons \& Fractals},
  volume={139},
  pages={110057},
  year={2020},
  publisher={Elsevier}
}

@article{murray2021potential,
  title={The potential future of the COVID-19 pandemic: will SARS-CoV-2 become a recurrent seasonal infection?},
  author={Murray, Christopher JL and Piot, Peter},
  journal={Jama},
  volume={325},
  number={13},
  pages={1249--1250},
  year={2021},
  publisher={American Medical Association}
}

@article{ndiaye2020analysis,
  title={Analysis of the COVID-19 pandemic by SIR model and machine learning technics for forecasting},
  author={Ndiaye, Babacar Mbaye and Tendeng, Lena and Seck, Diaraf},
  journal={arXiv preprint arXiv:2004.01574},
  year={2020}
}

@article{friedman2021predictive,
  title={Predictive performance of international COVID-19 mortality forecasting models},
  author={Friedman, Joseph and Liu, Patrick and Troeger, Christopher E and Carter, Austin and Reiner, Robert C and Barber, Ryan M and Collins, James and Lim, Stephen S and Pigott, David M and Vos, Theo and others},
  journal={Nature communications},
  volume={12},
  number={1},
  pages={1--13},
  year={2021},
  publisher={Nature Publishing Group}
}

@article{sahin2020weather,
title = {Impact of weather on COVID-19 pandemic in Turkey},
journal = {Science of The Total Environment},
volume = {728},
pages = {138810},
year = {2020},
issn = {0048-9697},
doi = {https://doi.org/10.1016/j.scitotenv.2020.138810},
url = {https://www.sciencedirect.com/science/article/pii/S0048969720323275},
author = {Mehmet Şahin}
}

@article{tosepu2020weather,
title = {Correlation between weather and Covid-19 pandemic in Jakarta, Indonesia},
journal = {Science of The Total Environment},
volume = {725},
pages = {138436},
year = {2020},
issn = {0048-9697},
doi = {https://doi.org/10.1016/j.scitotenv.2020.138436},
url = {https://www.sciencedirect.com/science/article/pii/S0048969720319495},
author = {Ramadhan Tosepu and Joko Gunawan and Devi Savitri Effendy and La Ode Ali Imran Ahmad and Hariati Lestari and Hartati Bahar and Pitrah Asfian},
}

@article{xi2020weather,
title = {Association between ambient temperature and COVID-19 infection in 122 cities from China},
journal = {Science of The Total Environment},
volume = {724},
pages = {138201},
year = {2020},
issn = {0048-9697},
doi = {https://doi.org/10.1016/j.scitotenv.2020.138201},
url = {https://www.sciencedirect.com/science/article/pii/S0048969720317149},
author = {Jingui Xie and Yongjian Zhu}
}

@article{ma2020weather,
title = {Effects of temperature variation and humidity on the death of COVID-19 in Wuhan, China},
journal = {Science of The Total Environment},
volume = {724},
pages = {138226},
year = {2020},
issn = {0048-9697},
doi = {https://doi.org/10.1016/j.scitotenv.2020.138226},
url = {https://www.sciencedirect.com/science/article/pii/S0048969720317393},
author = {Yueling Ma and Yadong Zhao and Jiangtao Liu and Xiaotao He and Bo Wang and Shihua Fu and Jun Yan and Jingping Niu and Ji Zhou and Bin Luo}
}

@article{rippinger2021evaluation,
  title={Evaluation of undetected cases during the COVID-19 epidemic in Austria},
  author={Rippinger, Claire and Bicher, Martin and Urach, Christoph and Brunmeir, Dominik and Weibrecht, N and Zauner, G and Sroczynski, G and Jahn, B and M{\"u}hlberger, N and Siebert, U and others},
  journal={BMC Infectious Diseases},
  volume={21},
  number={1},
  pages={1--11},
  year={2021},
  publisher={BioMed Central}
}

@article{jahn2021targeted,
  title={Targeted covid-19 vaccination (tav-covid) considering limited vaccination capacities—an agent-based modeling evaluation},
  author={Jahn, Beate and Sroczynski, Gaby and Bicher, Martin and Rippinger, Claire and M{\"u}hlberger, Nikolai and Santamaria, J{\'u}lia and Urach, Christoph and Schomaker, Michael and Stojkov, Igor and Schmid, Daniela and others},
  journal={Vaccines},
  volume={9},
  number={5},
  pages={434},
  year={2021},
  publisher={Multidisciplinary Digital Publishing Institute}
}

@article{bicher2021evaluation,
  title={Evaluation of Contact-Tracing Policies against the Spread of SARS-CoV-2 in Austria: An Agent-Based Simulation},
  author={Bicher, Martin and Rippinger, Claire and Urach, Christoph and Brunmeir, Dominik and Siebert, Uwe and Popper, Niki},
  journal={Medical Decision Making},
  pages={0272989X211013306},
  year={2021},
  publisher={SAGE Publications Sage CA: Los Angeles, CA}
}

@article{bicher2021model,
  title={Model Based Estimation of the SARS-CoV-2 Immunization Level in Austria and Consequences for Herd Immunity Effects},
  author={Bicher, Martin Richard and Rippinger, Claire and Schneckenreither, G{\"u}nter Richard and Weibrecht, Nadine and Urach, Christoph and Zechmeister, Melanie and Brunmeir, Dominik and Huf, Wolfgang and Popper, Niki},
  journal={medRxiv},
  year={2021},
  publisher={Cold Spring Harbor Laboratory Press}
}

@article{cov19law,
  author={PARLAMENT},
  title   = {COVID-19 Gesetz.},
  journal = {BGBl. I Nr. 12/2020},
  day={15},
  month={03},
  year={2020},
  url={https://www.ris.bka.gv.at/eli/bgbl/I/2020/12/20200315}
}

@article{cov19law2,
  author={BMSGPK},
  title   = {Vorläufige Maßnahmen zur Verhinderung der Verbreitung von COVID-19},
  journal = {BGBl. II Nr. 96/2020},
  day={15},
  month={03},
  year={2020},
  url = {https://www.ris.bka.gv.at/eli/bgbl/II/2020/96/20200315}
}

@article{cov19law3,
  author={BMSGPK},
  title   = {Verordnung, mit der zur Verhinderung der Verbreitung von COVID-19 die Sperrstunde und Aufsperrstunde im Gastgewerbe festgelegt werden},
  journal = {BGBl. II Nr. 96/2020},
  day={15},
  month={03},
  year={2020},
  url = {https://www.ris.bka.gv.at/eli/bgbl/II/2020/97/20200315}
}

@article{cov19law4,
  author={BMSGPK},
  title   = {Verordnung gemäß § 2 Z 1 des COVID-19-Maßnahmengesetzes},
  journal = {BGBl. II Nr. 98/2020},
  day={15},
  month={03},
  year={2020},
  url = {https://www.ris.bka.gv.at/eli/bgbl/II/2020/98/20200315}
}

@article{cov19law5,
  author={BMSGPK},
  title   = {Verordnung gemäß § 2 Z 1 des COVID-19-Maßnahmengesetzes},
  journal = {BGBl. II Nr. 107/2020},
  day={19},
  month={03},
  year={2020},
  url = {https://www.ris.bka.gv.at/eli/bgbl/II/2020/107/20200319}
}

@article{cov19law6,
  author={BMSGPK},
  title   = {Änderung der Verordnung betreffend vorläufige Maßnahmen zur Verhinderung der Verbreitung von COVID-19},
  journal = {BGBl. II Nr. 110/2020},
  day={20},
  month={03},
  year={2020},
  url = {https://www.ris.bka.gv.at/eli/bgbl/II/2020/110/20200320}
}

@article{cov19law7,
  author={PARLAMENT},
  title   = {2. COVID-19-Gesetz},
  journal = {BGBl. I Nr. 16/2020},
  day={21},
  month={03},
  year={2020},
  url = {https://www.ris.bka.gv.at/eli/bgbl/I/2020/16/20200321}
}

@article{cov19law8,
  author={BMSGPK},
  title   = {Änderung der Verordnung betreffend vorläufige Maßnahmen zur Verhinderung der Verbreitung von COVID-19},
  journal = {BGBl. II Nr. 112/2020},
  day={22},
  month={03},
  year={2020},
  url = {https://www.ris.bka.gv.at/eli/bgbl/II/2020/112/20200322}
}

@article{cov19law10,
  author={BMSGPK},
  title   = {2. COVID-19-Öffnungsverordnung und 1. Novelle zur 2. COVID-19-Öffnungsverordnung },
  journal = {BGBl. II Nr. 278/2021},
  day={28},
  month={06},
  year={2021},
  url = {https://www.ris.bka.gv.at/eli/bgbl/II/2021/278/20210628}
}

@article{cov19law11,
  author={BMSGPK},
  title   = {COVID-19-Öffnungsverordnung – COVID-19-ÖV und 1. Novelle zur COVID-19-Öffnungsverordnung},
  journal = {BGBl. II Nr. 214/2021},
  day={10},
  month={05},
  year={2021},
  url = {https://www.ris.bka.gv.at/eli/bgbl/II/2021/214/20210510}
}

@article{cov19law12,
  author={BMSGPK},
  title   = {2. Novelle zur COVID-19-Öffnungsverordnung},
  journal = {BGBl. II Nr. 223/2021},
  day={18},
  month={05},
  year={2021},
  url = {https://www.ris.bka.gv.at/eli/bgbl/II/2021/223/20210518}
}

@article{cov19law13,
  author={BMSGPK},
  title   = {3. Novelle zur COVID-19-Öffnungsverordnung},
  journal = {BGBl. II Nr. 242/2021},
  day={01},
  month={06},
  year={2021},
  url = {https://www.ris.bka.gv.at/eli/bgbl/II/2021/242/20210601}
}

@article{cov19law14,
  author={BMSGPK},
  title   = {4. Novelle zur COVID-19-Öffnungsverordnung},
  journal = {BGBl. II Nr. 247/2021},
  day={02},
  month={06},
  year={2021},
  url = {https://www.ris.bka.gv.at/eli/bgbl/II/2021/247/20210602}
}

@article{cov19law15,
  author={BMSGPK},
  title   = {5. Novelle zur COVID-19-Öffnungsverordnung},
  journal = {BGBl. II Nr. 256/2021},
  day={09},
  month={06},
  year={2021},
  url = {https://www.ris.bka.gv.at/eli/bgbl/II/2021/256/20210609}
}

@article{cov19law16,
  author={BMSGPK},
  title   = {Änderung der COVID-19-Einreiseverordnung},
  journal = {BGBl. II Nr. 220/2021},
  day={15},
  month={05},
  year={2021},
  url = {https://www.ris.bka.gv.at/eli/bgbl/II/2021/220/20210512}
}

@article{cov19law17,
  author={BMSGPK},
  title   = {COVID-19-Einreiseverordnung – COVID-19-EinreiseV 2021},
  journal = {BGBl. II Nr. 276/2021},
  day={25},
  month={06},
  year={2021},
  url = {https://www.ris.bka.gv.at/eli/bgbl/II/2021/276/20210625}
}

@article{cov19law18,
  author={PARLAMENT},
  title   = {3. COVID-19-Gesetz},
  journal = {BGBl. I Nr. 23/2020},
  day={04},
  month={04},
  year={2020},
  url = {https://www.ris.bka.gv.at/eli/bgbl/I/2020/23/20200404}
}

@article{cov19law19,
  author={PARLAMENT},
  title   = {4. COVID-19-Gesetz},
  journal = {BGBl. I Nr. 24/2020},
  day={04},
  month={04},
  year={2020},
  url = {https://www.ris.bka.gv.at/eli/bgbl/I/2020/24/20200404}
}

@article{cov19law20,
  author={PARLAMENT},
  title   = {5. COVID-19-Gesetz},
  journal = {BGBl. I Nr. 25/2020},
  day={04},
  month={04},
  year={2020},
  url = {https://www.ris.bka.gv.at/eli/bgbl/I/2020/25/20200404}
}

@misc{opendata21covid,
  title = {{BMSGPK} Österreichisches  covid-19  open  data  informationsportal,  covid-19:   Zeitverlauf  der  gemeldeteten  covid-19  zahlender bundesläander (morgenmeldung).},
  howpublished = {\url{https://www.data.gv.at/katalog/dataset/f33dc893-bd57-4e5c-a3b0-e32925f4f6b1}},
  note = {Accessed: 2021-03-14}
}

@misc{jhu21covid,
  author={Dong E., Du H. and Gardner, L.},
  title = {Covid-19 data repository by the center for systems science and engineering (csse) at johns hopkins university},
  howpublished = {\url{https://github.com/CSSEGISandData/COVID-19}},
  note = {Accessed: 2021-03-21}
}

@misc{bmsgpk21covidmodels,
  author={BMSGPK},
  title = {COVID-Prognose-Konsortium},
  howpublished = {\url{https://www.sozialministerium.at/Informationen-zum-Coronavirus/Neuartiges-Coronavirus-(2019-nCov)/COVID-Prognose-Konsortium.html}},
  note = {Accessed: 2021-07-03}
}

@inproceedings{kosiachenko2019mass,
  title={MASS CUDA: a general GPU parallelization framework for agent-based models},
  author={Kosiachenko, Lisa and Hart, Nathaniel and Fukuda, Munehiro},
  booktitle={International Conference on Practical Applications of Agents and Multi-Agent Systems},
  pages={139--152},
  year={2019},
  organization={Springer}
}

@article{hale2021global,
  title={A global panel database of pandemic policies (Oxford COVID-19 Government Response Tracker)},
  author={Hale, Thomas and Angrist, Noam and Goldszmidt, Rafael and Kira, Beatriz and Petherick, Anna and Phillips, Toby and Webster, Samuel and Cameron-Blake, Emily and Hallas, Laura and Majumdar, Saptarshi and others},
  journal={Nature Human Behaviour},
  volume={5},
  number={4},
  pages={529--538},
  year={2021},
  publisher={Nature Publishing Group}
}

@misc{hohewarte,
  title = {Monatliche Wetterdaten der Messstation Hohe Warte seit Jänner 1955 zu Lufttemperatur, Luftdruck, Bewölkung, Windgeschwindigkeit und Niederschlag. },
  author={MA23},
  howpublished = {\url{https://www.data.gv.at/katalog/dataset/wetter-seit-1955-hohe-warte-wien/}},
  note = {Accessed: 2021-07-12}
}

@misc{zamg,
  title = {Zentralanstalt für Meteorologie und Geodynamik},
  howpublished = {\url{https://www.zamg.ac.at/cms/de/aktuell}},
  note = {Accessed: 2021-07-12}
}

@misc{austat,
  title = {Bevölkerung nach Alter und Geschlecht},
  howpublished = {\url{https://www.statistik.at/web_de/statistiken/menschen_und_gesellschaft/bevoelkerung/bevoelkerungsstruktur/bevoelkerung_nach_alter_geschlecht/index.html}},
  note = {Accessed: 2021-07-12}
}

@misc{agesReff,
  title = {Epidemiologische Parameter des COVID19 Ausbruchs, Update 09.07.2021, Österreich, 2020/2021},
  author={Lukas Richter and Daniela Schmid, Martin Borkovec and Ali Chakeri, Sabine Maritschnik and Sabine Pfeiffer, Ernst Stadlober},
  howpublished = {\url{https://www.ages.at/wissen-aktuell/publikationen/epidemiologische-parameter-des-covid19-ausbruchs-oesterreich-20202021/}},
  note = {Accessed: 2021-07-12}
}

@misc{agesDash,
  title = {{AGES Dashboard COVID19} Datenstand des Epidemiologischen Meldesystems 13.07.2021},
  author={AGES},
  howpublished = {\url{https://covid19-dashboard.ages.at//}},
  note = {Accessed: 2021-07-13}
}

@Article{covid19datahub,
  title = {COVID-19 Data Hub},
  year = {2020},
  doi = {10.21105/joss.02376},
  author = {Emanuele Guidotti and David Ardia},
  journal = {Journal of Open Source Software},
  volume = {5},
  number = {51},
  pages = {2376},
}

@article{ripperger2020orthogonal,
  title={Orthogonal SARS-CoV-2 serological assays enable surveillance of low-prevalence communities and reveal durable humoral immunity},
  author={Ripperger, Tyler J and Uhrlaub, Jennifer L and Watanabe, Makiko and Wong, Rachel and Castaneda, Yvonne and Pizzato, Hannah A and Thompson, Mallory R and Bradshaw, Christine and Weinkauf, Craig C and Bime, Christian and others},
  journal={Immunity},
  volume={53},
  number={5},
  pages={925--933},
  year={2020},
  publisher={Elsevier}
}

@article{stuckler2009public,
  title={The public health effect of economic crises and alternative policy responses in Europe: an empirical analysis},
  author={Stuckler, David and Basu, Sanjay and Suhrcke, Marc and Coutts, Adam and McKee, Martin},
  journal={The Lancet},
  volume={374},
  number={9686},
  pages={315--323},
  year={2009},
  publisher={Elsevier}
}

@article{tapia2017population,
  title={Population health and the economy: Mortality and the Great Recession in Europe},
  author={Tapia Granados, Jos{\'e} A and Ionides, Edward L},
  journal={Health economics},
  volume={26},
  number={12},
  pages={e219--e235},
  year={2017},
  publisher={Wiley Online Library}
}

@article{stieger2020psychological,
  title={Psychological well-being under conditions of lockdown: an experience sampling study in Austria during the COVID-19 pandemic},
  author={Stieger, Stefan and Lewetz, David and Swami, Viren},
  year={2020},
  publisher={PsyArXiv}
}

@article{stieger2021emotional,
  title={Emotional well-being under conditions of lockdown: An experience sampling study in Austria during the COVID-19 pandemic},
  author={Stieger, Stefan and Lewetz, David and Swami, Viren},
  journal={Journal of happiness studies},
  pages={1--18},
  year={2021},
  publisher={Springer}
}

@article{ueda2020suicide,
  title={Suicide and mental health during the COVID-19 pandemic in Japan},
  author={Ueda, Michiko and Nordstr{\"o}m, Robert and Matsubayashi, Tetsuya},
  journal={Journal of public health (Oxford, England)},
  year={2020},
  publisher={Oxford University Press}
}

@article{stolz2021impact,
  title={The impact of COVID-19 restriction measures on loneliness among older adults in Austria},
  author={Stolz, Erwin and Mayerl, Hannes and Freidl, Wolfgang},
  journal={European journal of public health},
  volume={31},
  number={1},
  pages={44--49},
  year={2021},
  publisher={Oxford University Press}
}

@article{pieh2021mental,
  title={Mental health during COVID-19 lockdown in the United Kingdom},
  author={Pieh, Christoph and Budimir, Sanja and Delgadillo, Jaime and Barkham, Michael and Fontaine, Johnny RJ and Probst, Thomas},
  journal={Psychosomatic medicine},
  volume={83},
  number={4},
  pages={328--337},
  year={2021},
  publisher={LWW}
}

@article{pieh2021comparing,
  title={Comparing Mental Health During the COVID-19 Lockdown and 6 Months After the Lockdown in Austria: A Longitudinal Study},
  author={Pieh, Christoph and Budimir, Sanja and Humer, Elke and Probst, Thomas},
  journal={Frontiers in Psychiatry},
  volume={12},
  pages={197},
  year={2021},
  publisher={Frontiers}
}

@article{pieh2020effect,
  title={The effect of age, gender, income, work, and physical activity on mental health during coronavirus disease (COVID-19) lockdown in Austria},
  author={Pieh, Christoph and Budimir, Sanja and Probst, Thomas},
  journal={Journal of psychosomatic research},
  volume={136},
  pages={110186},
  year={2020},
  publisher={Elsevier}
}

@misc{nowotny2019depressionsbericht,
  title={Depressionsbericht {\"O}sterreich. Eine interdisziplin{\"a}re und multiperspektivische Bestandsaufnahme},
  author={Nowotny, Monika and Kern, Daniela and Breyer, Elisabeth and Bengough, Theresa and Griebler, Robert},
  year={2019},
  publisher={Bundesministerium f{\"u}r Arbeit, Soziales, Gesundheit und Konsumentenschutz}
}

@article{alboni2008there,
  title={Is there an association between depression and cardiovascular mortality or sudden death?},
  author={Alboni, Paolo and Favaron, Elisa and Paparella, Nelly and Sciammarella, Massimo and Pedaci, Mario},
  journal={Journal of Cardiovascular Medicine},
  volume={9},
  number={4},
  pages={356--362},
  year={2008},
  publisher={LWW}
}

@article{radeloff2021trends,
  title={Trends in suicide rates during the COVID-19 pandemic restrictions in a major German city},
  author={Radeloff, Daniel and Papsdorf, Rainer and Uhlig, Kirsten and Vasilache, Andreas and Putnam, Karen and Von Klitzing, Kai},
  journal={Epidemiology and psychiatric sciences},
  volume={30},
  year={2021},
  publisher={Cambridge University Press}
}

@article{ruck2020will,
  title={Will the COVID-19 pandemic lead to a tsunami of suicides? A Swedish nationwide analysis of historical and 2020 data},
  author={Ruck, Christian and Mataix-Cols, David and Malki, Kinda and Adler, Mats and Flygare, Oskar and Runeson, Bo and Sidorchuk, Anna},
  journal={MedRxiv},
  year={2020},
  publisher={Cold Spring Harbor Laboratory Press}
}

@article{reeves2015economic,
  title={Economic shocks, resilience, and male suicides in the Great Recession: cross-national analysis of 20 EU countries},
  author={Reeves, Aaron and McKee, Martin and Gunnell, David and Chang, Shu-Sen and Basu, Sanjay and Barr, Benjamin and Stuckler, David},
  journal={The European Journal of Public Health},
  volume={25},
  number={3},
  pages={404--409},
  year={2015},
  publisher={Oxford University Press}
}

@article{jackson2013depression,
  title={Depression and risk of stroke in midaged women: a prospective longitudinal study},
  author={Jackson, Caroline A and Mishra, Gita D},
  journal={Stroke},
  volume={44},
  number={6},
  pages={1555--1560},
  year={2013},
  publisher={Am Heart Assoc}
}

@article{metzler2020decline,
  title={Decline of acute coronary syndrome admissions in Austria since the outbreak of COVID-19: the pandemic response causes cardiac collateral damage},
  author={Metzler, Bernhard and Siostrzonek, Peter and Binder, Ronald K and Bauer, Axel and Reinstadler, Sebastian Johannes},
  journal={European heart journal},
  volume={41},
  number={19},
  pages={1852--1853},
  year={2020},
  publisher={Oxford University Press}
}

@article{freudenberg2020impact,
  title={Impact of COVID-19 on nuclear medicine in Germany, Austria and Switzerland: an international survey in April 2020},
  author={Freudenberg, Lutz S and Dittmer, Ulf and Herrmann, Ken},
  journal={Nuklearmedizin},
  volume={59},
  number={04},
  pages={294--299},
  year={2020},
  publisher={{\copyright} Georg Thieme Verlag KG}
}

@article{negrini2020up,
  title={Up to 2.2 million people experiencing disability suffer collateral damage each day of COVID-19 lockdown in Europe.},
  author={Negrini, Stefano and Grabljevec, Klemen and Boldrini, Paolo and Kiekens, Carlotte and Moslavac, Sasa and Zampolini, Mauro and Christodoulou, Nicolas},
  journal={Eur J Phys Rehabil Med},
  pages={361--365},
  year={2020}
}

@article{steiber2021covid,
  title={Die COVID-19 Gesundheits-und Arbeitsmarktkrise und ihre Auswirkungen auf die Bev{\"o}lkerung},
  author={Steiber, Nadia},
  journal={Materialien zu Wirtschaft und Gesellschaft. Working Paper-Reihe der AK Wien},
  number={211},
  year={2021},
  publisher={AK Wien}
}

@article{holzl2021zweite,
  title={Zweite COVID-19-Welle bestimmt Konjunkturbeurteilung der Unternehmen. Ergebnisse der Quartalsbefragung des WIFO-Konjunkturtests vom J{\"a}nner 2021},
  author={H{\"o}lzl, Werner and Klien, Michael and K{\"u}gler, Agnes and others},
  journal={WIFO Monatsberichte (monthly reports)},
  volume={94},
  number={2},
  pages={105--114},
  year={2021},
  publisher={WIFO}
}

@misc{oecd,
  title = {OECD Data},
  howpublished = {\url{https://data.oecd.org/}},
  note = {Accessed: 2021-03-24}
}

\end{document}